%% file: main.tex
\documentclass[]{interact}

\usepackage[T1]{fontenc}
\usepackage[utf8]{inputenc}
\usepackage{xcolor}

\usepackage[
  inline
]{enumitem}

\usepackage{
  amsmath,
  amssymb,
  amsfonts,
  booktabs,
  cite,
  framed,
  mathtools,
  microtype,
  multirow,
  nicefrac,
  subcaption,
  tabularx,
  url,
  listings,
  xcolor
}

\usepackage[
  pdftex,
  colorlinks = true,
  bookmarks = false,
  linkcolor = blue,
  citecolor = blue,
  urlcolor = blue,
  bookmarks
]{hyperref}

\usepackage[
  capitalize
]{cleveref}

\usepackage[
  scaled = 0.82
]{helvet}

\usepackage[
  scaled=0.80
]{beramono}

\definecolor{darkblue}{rgb}{0.0,0.0,0.6}
\definecolor{darkred}{rgb}{0.6,0.1,0.1}

\lstdefinelanguage{Xtext}{
  morestring=[b]',
  stringstyle=\color{darkred},
  keywordstyle=\color{darkblue},
  morekeywords={Artifact, Node, Publisher, RosPublisher, ComponentInterface, RosSystem, RosSubscriber, RefPublisher, RefSubscriber, TopicConnection, From, To}
}

\lstdefinestyle{XtextStyle}{
  language=Xtext,
  tabsize=2,
  showspaces=false,
  showstringspaces=false
}

\xdefinecolor{DarkGreenGray}{HTML}{697060} 


\newcommand{\tomasys}[1]{\textsf{#1}}

\begin{document}

\title{MROS:\ Runtime Adaptation\\For Robot Control Architectures}

\author{Darko Bozhinoski, Mario Garzon Oviedo, Nadia Hammoudeh Garcia, Harshavardhan Deshpande, Gijs van der Hoorn, Jon Tjerngren, Andrzej Wąsowski and Carlos Hernandez Corbato}



\maketitle

\thispagestyle{plain}
\pagestyle{plain}

\begin{abstract}

  Known attempts to build autonomous robots rely on complex control architectures, usually implemented with the Robot Operating System (ROS).  Runtime adaptation is needed in these systems, to cope with component failures and with contingencies arising from dynamic environments---otherwise these affect the reliability and quality of the mission execution. Existing proposals on how to build self-adaptive systems in robotics usually require a major re-design of the control architecture and rely on complex tools unfamiliar to the robotics community.  Moreover they are hard to reuse across applications.

  This paper present MROS: a model-based framework for run-time adaptation of robot control architectures based on ROS.\@ MROS uses a combination of domain specific languages to model architectural variants and capture mission quality concerns, and an ontology-based implementation of the MAPE-K and meta-control visions for runtime adaptation.  The experiment results obtained applying MROS in two realistic ROS-based robotic demonstrators show the benefits of our approach in terms of the quality of the mission execution, and MROS's extensibility and re-usability across robotic applications.

\end{abstract}

\maketitle


\input{introduction.tex}
\input{background.tex}
\input{solution.tex}

\input{demonstrators.tex}

\input{quantitativeevaluation.tex}

\input{qualitativeevaluation.tex}
\input{related.tex}
\input{conclusion.tex}

\bibliographystyle{IEEEtranS}
\bibliography{references}

\end{document}

%% file: introduction.tex
\section{Introduction}%
\label{sec:introduction}

\noindent
Robotic systems are designed by integrating and configuring individual robot capabilities in a governing control architecture, most frequently based on the Robot Operating System (ROS)\,\cite{Quigley-2009}. 
However, static control architectures fall short when addressing context variability in open-ended, dynamic environments, where internal errors also compromise the quality and autonomy of mission execution. 
 Self-adaptive Systems methods offer various solutions to deal with the context variability resulting from uncertainty and contingencies during execution\cite{Weyns-2017}.  However, existing approaches to apply these methods in robotic systems are marginal, because they require a major re-design of the control architecture, they use complex tools unfamiliar to the robotics community, and they are hard to reuse across applications.  Therefore, there is a pressing demand for solutions to integrate self-adaptation in robot control architectures.  This paper demonstrates how model-based self-adaptation can be easily integrated in robot control architectures to increase mission reliability and quality.
\looseness=-1

Robotics researchers have come up with sophisticated control architectures that are able to perform very well a specific mission~\cite{Thrun-2006, Correll-2016}. However, these architectures are mission and robot platform specific and are unable to address other missions and robots without undergoing major modifications~\cite{Hernandez-2018}.

\begin{figure}[t]

  \begin{center}
    \includegraphics[
      width = \linewidth,
      clip,
      trim  = 5mm 0mm 0mm 0mm
    ]{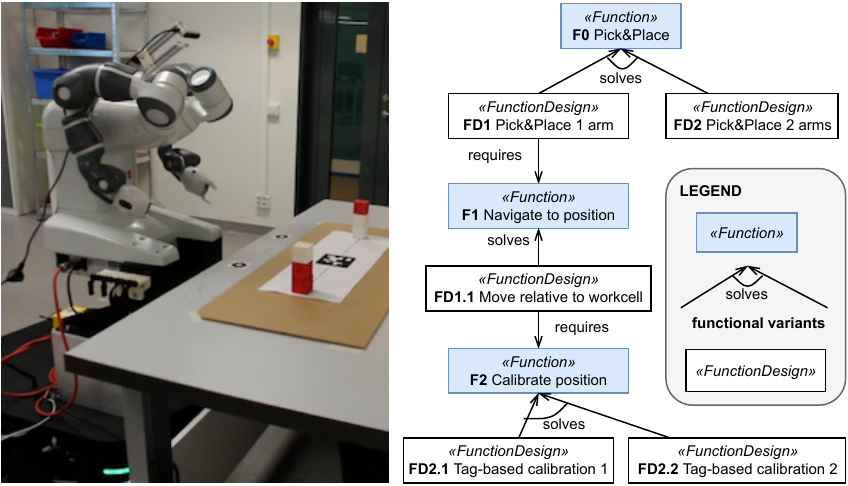}

  \end{center}

  \vspace{-1mm}

  \caption{Left: Demonstrator D1 based on an ABB YuMi manipulator. Right: the TOMASys model of its control architecture.}%
  \label{fig:arch-model}
  \label{fig:pyramid}


  \vspace{-2mm plus 0.5mm minus 0.5mm}

\end{figure}

An example of a robot performing a complex task is the mobile manipulator shown on  \Cref{fig:arch-model}. The robot uses QR-tags to calibrate its position in the work cell, and the wooden blocks with which it builds a pyramid.
The original ROS-based control architecture suffered from errors locating the QR-tag in varying light conditions, and from task failures whenever an arm bumps lightly into any obstacle (the internal safety function prevents any further motion of that arm).  However, the system offers redundancy that can be exploited for self-adaptation.
The challenge is to integrate self-adaptation into its architecture without major development effort, and maintaining extensibility to other tasks (e.g. navigation between the workstations), so that it can be reused for other robots.

We claim that using a model-based solution to integrate self-adaptation in robot control architectures addresses this challenge and results in increased mission quality and reliability at a lower cost.
Modeling has been successfully applied in various domains to solve issues of architectural composition both statically (software product lines~\cite{pohl.ea:2005}) and at runtime (dynamic software product lines~\cite{DBLP:journals/jss/CapillaBTCH14}, models-at-runtime~\cite{bencomo14dagstuhl,Blair-2009}).
Using models of the robot control architecture raises the abstraction level in the implementation of robot control systems, improving reuse from system to system, which is presently rare and often \emph{ad hoc}.
While work has been done in this direction \cite{Brugali-2018}, given the complexity of robotics systems, it is unclear how to proceed to maximize the benefits, and to avoid redesigning and re-implementing the adaptation layer for each robot from scratch.
\looseness=-1


In this work, we propose MROS:
a model-based approach for self-adaptation of robot control architectures, contributing two key insights. First, the use of different Domain Specific Languages to model different levels of abstraction allows for lean and extensible tools to design self-adaptation in robotics. Second, an explicit separation of concerns, between mission-specific logic and system management, and between general and mission-specific concerns, allows to reuse adaptation rules and artifacts across missions and robotic systems.
For system design, MROS uses a combination of platform-specific Domain Specific Languages  (DSLs) to model the robot architecture. These DSLs are close to the ROS ecosystem and 
effectively capture functional and task quality concerns of a robotic system for run-time adaptation.
For run time adaptation, MROS realizes the \emph{meta-control} vision\cite{Hernandez-2018a} using a novel implementation of the MAPE-K loop based on ontological reasoning to drive the run-time adaptations.
\looseness=-1

A parallel, but equally important, goal of this paper is to advance experiment design for model-based self-adaptation, to push the community from anecdotal evidence towards data.  From methodological perspective, we propose to evaluate self-adaptation frameworks at two levels: (i) how well, and at what computational cost, a given framework improves mission performance and reliability, and (ii) how reusable (vs idiosyncratic) a framework is; to how large class of systems it applies, and how much specialized development is needed.  We evaluate the former in a quantitative experiment, and the latter in a qualitative assessment based on engineering experience from the project, and its architectural qualities.


Our main contributions include:

\begin{itemize}

  \item \emph{A reusable implementation of MROS meta-controller,} ready to integrate self-adaptation in ROS systems.

  \item \emph{A set of design tools and languages} to model and deploy the MROS solution in ROS systems.

  \item \emph{An evaluation of MROS in two case studies (mobile manipulator, factory floor navigation)} exceeding prior methodological standards.  The experiments show performance and robustness improvements, and quantify the computational cost of the general architecture.

  \item \emph{An analysis of benefits of model-driven methods for this problem.}

\end{itemize}

\noindent Design methodologies for autonomy and self-adaptation remain an open discussion topic in the robotics community.  With this paper, developed in a joint effort of robotics and modeling researchers, we hope to get more members in the modeling community interested in the problem. We also contribute a validated approach and artifacts to the discussion of systematic design methods and reusable architectural components for robotics.  We believe that the integration of Meta-control with ROS may bring models-at-runtime and system modeling methods to broader groups of practitioners in robotics.

The rest of the paper is organized as follows.
\Cref{sec:background} gives background information for model-based development and ROS, and self-adaptation using meta-control.
\Cref{sec:demo} presents the two robotic demonstrators developed to validate our solution. \Cref{sec:quantity_evaluation} defines our experimental setup and results regarding the run time feasibility of MROS on both demonstrators.  \Cref{sec:quality_evaluation} discusses how MROS  supports system extensibility and reusability,
\Cref{sec:related} discusses related work and, \cref{sec:conclusion} presents the conclusions and future research direction.


%% file: background.tex
\section{Background}\label{sec:background}

\subsection{ROS and the RosModel tooling}

\begin{figure}[t]

  \begin{center}
    \includegraphics[
      width = 0.8 \columnwidth,
    ]{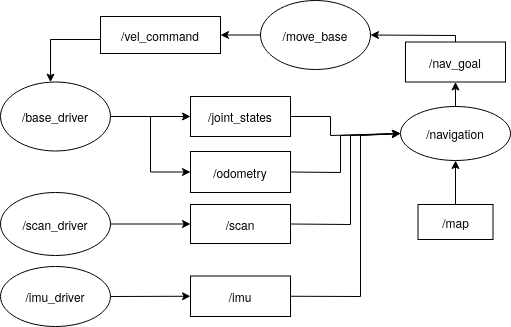}
  \end{center}

  \vspace{-2.2mm plus 0.5mm}

  \caption{ROS graph for a simple navigation application, extracted from a system at runtime by standard ROS tools. Ellipses denote nodes, boxes mark topics.\looseness=-1}
  \label{fig:rosgraph}

  \vspace{-3mm plus 1mm}

\end{figure}

\noindent
The Robot Operating System is a component-based, an open-source platform considered the \emph{de facto} standard for the development of robotics systems, and the platform chosen to develop this work.  A key to the success of ROS lies in imposing few architectural constraints and in offering many tailored components (robotics-specific functional components and hardware drivers).  A typical ROS system consists of many small and mostly independent distributed programs called \textit{nodes}. Nodes communicate via message passing, using a publish-subscribe mechanism, or service calls.  \Cref{fig:rosgraph} shows an example of a runtime communication graph of one of the cases of this paper.
\looseness=-1

\begin{figure}[t]

  \begin{center}
    \includegraphics[
      width = .94 \columnwidth,
      trim  = 0mm 13mm 0mm 31mm,
      clip
    ]{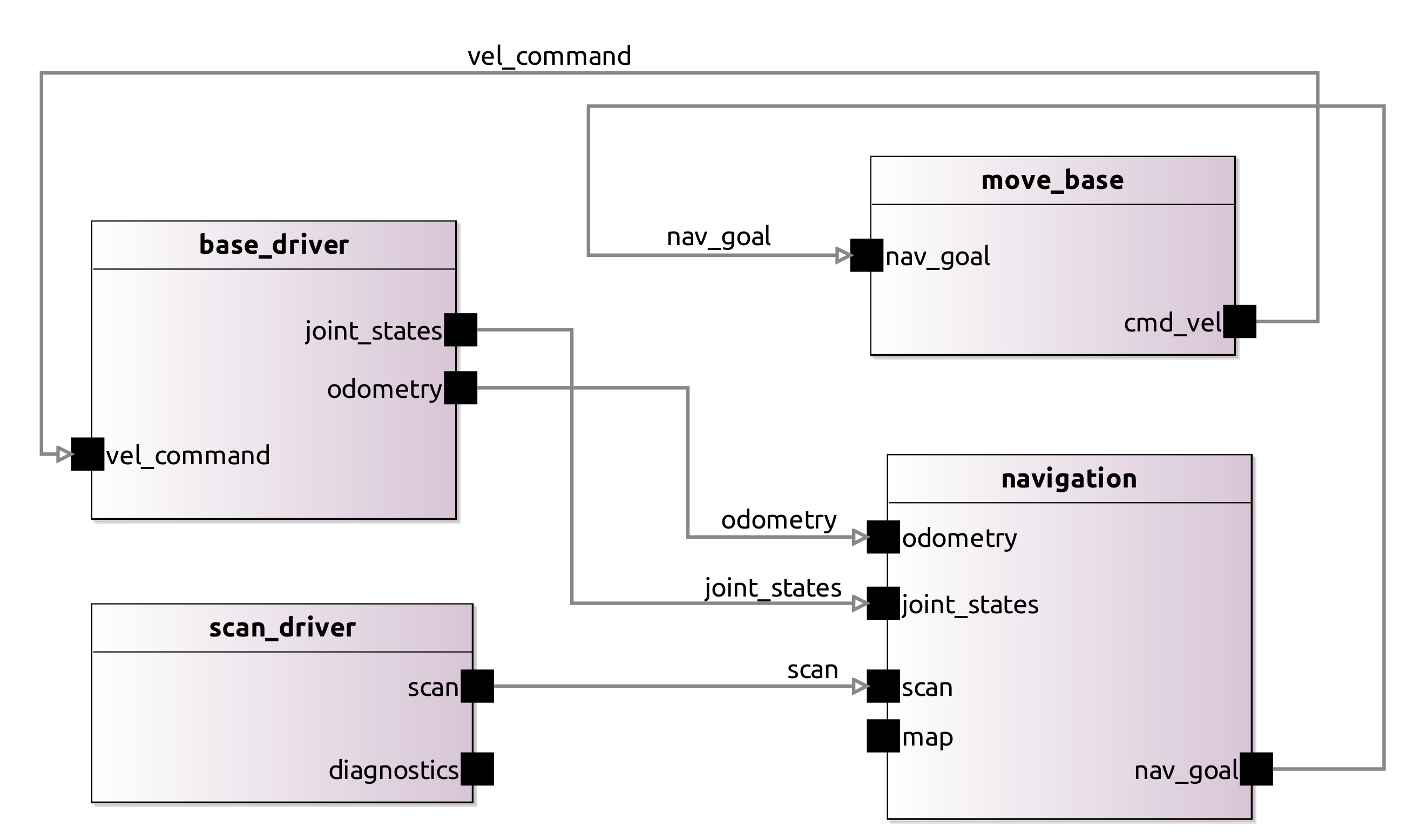}
  \end{center}

  \vspace{-2mm plus 0.5mm}

  \caption{Example RosSystem model for the system in \cref{fig:rosgraph}}%
  \label{fig:rossystem_capture}
  \vspace{-3mm}

\end{figure}

To facilitate model-based development with ROS, Hammoudeh Garcia and colleagues
developed the \textit{RosModel tooling} framework which contains a set of languages to formalize relevant properties of a ROS system architecture, in a manner compatible with model-based ecosystem \cite{Hammoudeh-2019}.  Our paper uses two of these languages: 
\begin{itemize}
    \item \emph{RosModel} model: specifies the communication interfaces of each node (the ports offered by a node) and the deployment information (the ROS package of a node and the packages it depends on)
    \item \textit{RosSystem} model: reflects the composition of the instantiated nodes, how they are wired and configured.
\end{itemize}
 \Cref{fig:rossystem_capture} shows the (static) \emph{RosSystem} for the (runtime) example of \cref{fig:rosgraph}, in the \textit{RosModel tooling} graphical editor.  Nodes are defined as components that contain a set of interfaces, with properties \textit{name} and a reference to the definition of the instantiated interface. The format also captures connections between the interfaces. Associated tools validate all the connections (i.e.\ the match of the type of the message sent by the publisher and expected by the subscriber) and automatically generate a \textit{launch file}, an XML script used in ROS to start, configure, and connect all the nodes that constitute the robotics system.  The tooling offers static analyzers and runtime introspectors that allow scaffolding the models from an existing software implementation.
\looseness=-1



\subsection{Meta-control and the TOMASys Meta-model}

\noindent
Meta-control\,\cite{Hernandez-2018a, Aguado-2021} is a reference architecture for self-adaptive autonomous control systems inspired by \hbox{MAPE-K} (\emph{Monitor-Analyze-Plan-Execute over a Knowledge base}, \cite{Kephart-2003a}) and supervisory control \cite{Blanke-2006}. At its center is the \emph{Teleological and Ontological Model for Autonomous Systems} (TOMASys) that allows to model the architecture of component-based systems, including architectural variations at both design-time and run-time \cite{Hernandez-2018a}. Architectural models provide the appropriate level of abstraction to manage adaptation at runtime \cite{Garlan-2014}. TOMASys captures the relation between the system's requirements and their allocation to the system components through the concept of functional and physical architectures from systems engineering \cite{Fernandez-2019,Hernandez-2017a}. In the Meta-control vision, a Meta-controller component implements runtime adaptation using the TOMASys model of the system and automatic reasoning for the evaluation, selection, and deployment of architectural variants.
\looseness=-1

In TOMASys, a \emph{function} represents a capability that has been designed in the system, for example ``Navigation'' in a mobile robot. A \emph{function design} is a possible realization of a function, an architectural variant able to deliver it.  Internally, a function design prescribes a certain structure that delivers the function, i.e.\ it maps functionally to system structure, through specifications of components and their interconnection.  TOMASys allows to model functional decomposition by designating \emph{required} functions for a function design, capturing the dependencies between lower- and higher-level functionality.  At runtime, the functional requirements are represented as instances called \textit{objectives}, which are solved by instances of the \textit{function designs}, the active function designs called \textit{function groundings}.
\looseness=-1

A TOMASys model example is shown in Fig.\,\ref{fig:pyramid}, using UML notation with stereotypes.  A high-level function \tomasys{F0} (Pick\&Place) represents the capability of the ABB mobile manipulator to pick and place small objects.  The model defines two function designs named \tomasys{FD1} and \tomasys{FD2} that \tomasys{solve} this function. \tomasys{FD2} represents an architectural variant that solves Pick\&Place using both arms of the robot (which jointly can reach the entire workspace), and it is implemented as a monolithic program in the ABB programming language. \tomasys{FD1} represents a variant that solves the function using only one arm and the ability of the mobile base to move and approach objects if they are out of reach of the arm.  It is implemented as a set of ROS nodes specifically parametrized to solve \tomasys{F0}, and it depends on function \tomasys{F1} (Move position) implemented by function design \tomasys{FD1.1} (Move relative to work cell). \tomasys{FD1.1} in turn requires the function \tomasys{F2} (Calibrate position) delivered by two designs, \tomasys{FD2.1} and \tomasys{FD2.2}, which detect the QR tag to calibrate the position of the robot in the work cell. \tomasys{FD2.1} and \tomasys{FD2.2} use different camera settings configured for different light conditions.
\looseness=-1

%% file: solution.tex
\begin{figure}[t]


  \includegraphics[width=\columnwidth]{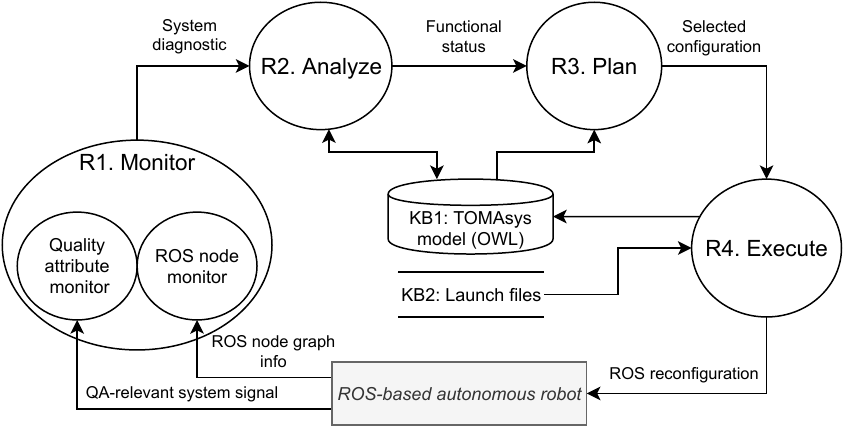}

  \vspace{-1mm}

  \caption{Data flow diagram for the run-time operation of the MAPE-K loop implemented by the MROS Meta-controller.}%
  \label{fig:mapek}

  \vspace{-3mm}

\end{figure}








\section{MROS Model-Based Meta-control}%
\label{sec:solution}

\noindent
MROS is a model-based solution for meta-control that integrates easily in a typical ROS development workflow.  It provides a general reusable and extensible meta-control node that extends any ROS system with runtime self-adaptation.  Reuse is maximized by separating the general reasoning principles, the adaptation related to system configuration, and error handling from task-related adaptations. Extensibility is achieved by using ontological reasoning to implement the decision-making in the meta-controller.  At design time, MROS greatly simplifies the development effort by extending the \textit{RosModel tooling} with a series of plugins that allow ROS developers to model architectural variants and their quality attributes, and to automate the generation and deployment of the meta-controller.
\looseness=-1

\subsection{Runtime Adaptation}


\noindent
MROS implements a MAPE-K loop using a runtime model conforming to the TOMASys meta-model serving as a knowledge base (KB1).  The realization of MAPE-K with ontological reasoning is a core contribution of this work. It allows to separate:
\begin{enumerate*}[label=(\roman*)]

  \item the TOMASys model of the robotic system, coded with the Ontology Web Language (OWL),

  \item the adaptation rules, coded with the Semantic Web Rule Language (SWRL), and

  \item the self-adaptation process---a ROS node \texttt{mros1\_reasoner}.

\end{enumerate*}
\Cref{fig:mapek} summarizes the main data processing steps in the MROS MAPE-K loop.  We discuss them in order below.

\subsubsection*{Step R1. Monitor}

Two monitoring processes track the status of active ROS nodes and the degree of fulfillment of non-functional requirements using quality attributes as a proxy.  The goal is to
\begin{enumerate*}[label=(\roman*)]

  \item detect and identify parameters misconfigured by developers in ways that lead to sub-optimal robot behavior,

  \item identify erroneous node behavior, and

  \item observe violation of system constraints (i.e. level of quality attributes).

\end{enumerate*}
The monitors operate at a fixed frequency and produce diagnostics using the standard ROS diagnostic mechanism \cite{diagnostics}. This means they easily integrate into an existing system, and might use or provide services already needed in the system, regardless of the presence of a meta-controller.

The \emph{ROS node monitors} leverage \emph{RosModel} tools. The existing \texttt{ros\_graph\_parser} node is used to create a \emph{RosSystem} model of the running ROS-based system. A newly developed \texttt{rosgraph\_monitor} compares that model with the desired \emph{RosSystem} model of base application (cf.\,\cref{tab:steps}). It publishes a diagnostic error message in case a difference is detected.  The  \emph{Quality attribute monitors} are implemented by a separate \emph{observer} node per each quality attribute.  They report normalized values for the quality attributes: zero means that the quality attribute is at the required level, one denotes maximum deviation from the required level. The skeleton of these nodes is automatically created from the \emph{RosSystem} model, for the attributes and the system signals (ROS topics) specified by the system architect.  Developers need to implement the logic in the observers to obtain the quality attribute level from the given signals.  For the evaluation cases presented below, we implemented observers for two attributes: safety (risk of collision with humans, using proximity sensors) and energy (using the runtime battery consumption and a battery model).
\looseness=-1

\subsubsection*{Step R2. Analyze}

The system diagnostic data produced in Step R1 is used to infer the status of the functional architecture of the system.  Any new facts, for instance a new component status when a node has reported a fatal error, or a change in the level of a quality attribute, are asserted in the TOMASys model (KB1). Then, ontological reasoning and application-independent rules (declared in SWRL, as proposed in \cite{Hernandez-2020}) are applied to the model to infer the current status of the functional architecture, including the status of the TOMASys objectives and the applicability of the alternative function designs. Finally, the status is updated in KB1.
\looseness=-1

\subsubsection*{Step R3. Plan}

If the meta-controller discovers in Step R2 that any objectives are violated, an architectural adaptation is proposed by searching for alternative function designs in the TOMASys model (KB1). If several available designs meet the required quality attribute levels, the one maximizing an application-specific utility function is selected. The \texttt{mros1\_reasoner}, used in R2--R3, is implemented with the Owlready2 library \cite{Lamy-2017}, the off-the-shelf reasoner Pellet to infer the functional status, and a custom logic for the search and selection of system configuration in the Plan step.
\looseness=-1

\subsubsection*{Step R4. Execute}

Enforce the adaptation selected in Step R3---stop nodes no longer needed, start new nodes needed, reconfigure the nodes remaining.  It is implemented by a general \texttt{rosgraph\textunderscore manipulator}, which follows an adaptation tactic suitable for ROS1, re-deploying the nodes by using launch files created by the \textit{RosModel} tools.

\looseness=-1

\subsection{Design Time Activities}





\noindent
The application developer uses the MROS extension of the \emph{RosModel} tools to describe the architecture of the domain application by specifying the set of possible configurations of components and their quality attributes.  \Cref{tab:steps}
summarizes the steps involved.

\input{design_steps_table.tex}








\begin{figure*}[t]

  \centering

  \begin{tabular}{
    c
    c
    >{\hspace{2mm}}c
  }

    \begin{lstlisting}
RosSystem { Name 'system_a'  RosComponents (
 ComponentInterface { name move_base
  RosParameters{
   RosParameter 'max_vel_x' { value 0.5 },
   RosParameter 'max_vel_y' { value 0.5 },
   RosParameter 'inflation_radius' { value 0.5 },
   RosParameter 'observation_sources' { value scan }}})
 Parameters {
  Parameter { name 'qa_safety' type Double value 0.41 },
  Parameter { name 'qa_energy' type Double value 0.48 }}}
    \end{lstlisting}

   & \hspace{5mm} &

    \begin{lstlisting}
RosSystem { Name 'system_b'  RosComponents (
 ComponentInterface { name move_base
  RosParameters{
   RosParameter 'max_vel_x' { value 0.3 },
   RosParameter 'max_vel_y' { value 0.3 },
   RosParameter 'inflation_radius' { value 0.8 },
   RosParameter 'observation_sources' { value point_cloud }}})
 Parameters {
  Parameter { name 'qa_safety' type Double value 0.71 },
  Parameter { name 'qa_energy' type Double value 0.33 }}}
    \end{lstlisting}

  \end{tabular}

  \caption{Fragments of \emph{RosSystem} models for two variants of a navigation system, specifying configurations of the \texttt{mode\_base} node and expected values for the quality attributes.  Left: base variant using a laser scanner, Right: alternative using a camera.}%
  \label{fig:rossystem}

  \vspace{-2mm plus 0.5mm minus 0.5mm}

\end{figure*}

\subsubsection*{Step 1. Model the application}

We first create a \textit{RosSystem} model of the base application, starting with individual nodes.  Two tools for automatic model generation from source code are available from the \emph{RosModel} project \cite{Hammoudeh-2021}: a built-in plugin that generates the models for ROS packages stored locally, and a web interface, available at \url{http://ros-model.seronet-project.de/}, for ROS packages publicly available.  Once the \emph{ROSModel} models describing single nodes are obtained, the developer can compose them, using either the graphical or the textual editor, to define the \textit{RosSystem} model of the entire base application.  Optionally, if MROS is used for an existing ROS application, the \textit{RosModel} tools support automatic analysis of the complete existing system.  This step results in a \textit{RosSystem} model (a \texttt{.rossystem} file) which is validated by our tools and used to automatically generate executable artifacts (launch files to start the ROS system).

\subsubsection*{Step 2. Model architectural variants}

Next, the developer models the adaptation possibilities---the architectural variants.  Most adaptations can be sufficiently captured by parameter variations, but larger architectural variations are also possible like switching to a different hardware module).  
\Cref{fig:rossystem} shows the representation of two different configurations of the same system. In this case, we have two different configurations of the node  \texttt{move\_base}. This node is in charge of interpreting the actions from the navigation module and translating them into commands to the base. Notice, how the maximum speed, inflation radius, and the type of sensor data used to adapt to the environment varies. The base variant \texttt{system\_a} uses the output of a laser proximity scanner, while the alternative variant \texttt{system\_b} uses the point cloud obtained from a 3D camera.  Additionally global parameters are added to define the value of the quality attributes (abbreviated \texttt{qa}), in the case of this example for safety and energy.

\subsubsection*{Step 3. Generate meta-controller deployment configuration}

We provide a \emph{RosModel} plugin that automatically generates the application-specific configuration for the \texttt{mros1\_reasoner} node to be integrated on top of the base application.
\looseness=-1

\subsubsection*{Step 4. Generate deployment configurations for variants}

We automatically generate executable artifacts (KB2: ROS launch files in \Cref{fig:mapek}), which can be called to execute runtime adaptations when a new configuration is selected by the reasoner (R4 in \Cref{fig:mapek}).

\subsubsection*{Step 5. Generate observers of quality attributes}

Another \textit{RosModel} wizard is used to generate the scripts to observe the system. The GUI allows to select the quality attributes to be monitored. MROS generates a description of the ROS node as a \textit{RosModel} and the skeleton of the Python code for monitoring.
\looseness=-1

\subsubsection*{Step 6. Generate runtime model for the meta-controller}

Finally, MROS provides a script that given the \textit{RosSystem} models of different adaptation strategies, automatically generates the TOMASys model that feeds the meta-controller at runtime.

In summary, the provided tools offer a holistic model-driven solution for the adaptation problem, exploiting models and automation, with a high degree of reuse both at design time \emph{and} at runtime. Since many artifacts are needed 
, generating all of them offers not only productivity gains: the tools ensure consistency of the input artifacts and guarantee consistency of all the generated artifacts.
\looseness=-1

%

%

%


%% file: design_steps_table.tex













\begin{table}[t]

  \renewcommand{\arraystretch}{1.00}
  \setlength{\tabcolsep}{3pt}

  \begin{tabularx}{\linewidth}{
      >{\hspace{.5pt}\raggedleft\bfseries}l
      >{\raggedright}p{31mm}
      >{\raggedright}X
      >{\raggedright\arraybackslash}p{31mm}
  }
    \emph{Step}
    & \emph{Input artifact}
    & \emph{Design-time activity}
    & \emph{Output artifact}
    \\ \toprule

    1
    & Application requirements
    & Application modeling (manual + reverse engineering tools)
    & \emph{RosSystem}\,model of\,base\,application
    \\

    2
    & \emph{RosSystem}\,model of\,base\,application
    & Modeling architectural variants
    & \emph{RosSystem}\,models
    \\

    3
    & \emph{RosSystem}\,model of\,base\,application
    & Generate meta-controller deployment configuration
    & (YAML)
    \\

    4
    & \emph{RosSystem}\,models
    & Generate deployment con\-fi\-gu\-ra\-tion\looseness=-1\ for variants (\textbf{automatic})
    & ROS Launch files
    \\

    5
    & \emph{RosSystem}\,models
    & Generate observers of the quality attributes (\textbf{automatic})
    & Observers boilerplate code
    \\

    6
    & \emph{RosSystem}\,models
    & Generate runtime model for the meta-controller (\textbf{automatic})
    & \emph{TOMASys model (OWL)}
    \\ \bottomrule

  \end{tabularx}

  \caption{Key steps in development of MROS meta-controller.}%
  \label{tab:steps}

  \vspace{-2mm plus 0.5mm minus 0.5mm}

\end{table}

%% file: demonstrators.tex
\section {MROS Demonstrators}%
\label{sec:demo}

\noindent
We use two different robotic systems to validate MROS and the developed artifacts.\footnote{The source code of the meta-controller and the experiment scripts is released under the open source license and can be found in the following repos: https://github.com/rosin-project/metacontrol\_sim (Metacontrol) and https://github.com/rosin-project/metacontrol\_experiments (experimental scripts)}
Both are
based on realistic applications for autonomous robots, concretely manipulation and navigation\,\cite{ABB-2019}. Mobile manipulation robots are designed to operate in semi-structured environments, such as a warehouse, a factory floor, and a healthcare facility. These environments change dynamically (e.g. the floor plan can change from day to day, some areas may be shared with humans and with other robots, etc). This context variability demands varying configurations of the robot
control architecture and appropriate methods to support successful mission completion.

\subsection{Dual-Arm Manipulator (D1)}\label{sec:yumi}

\noindent

In D1, the robot is a YuMi dual-arm manipulator standing on a Clearpath Ridgback mobile platform and equipped with an Intel Realsense RGBD camera (\cref{fig:pyramid}). Its task is to build a pyramid with wooden blocks at the work station.
The camera and QR-tags are used to calibrate the robot's position in the work cell, and to locate the bricks and the target position of the pyramid.
This information and prior knowledge about the blocks sizes, is used to compute the required manipulation motions.
To improve robustness, we use a YuMi dual-arm manipulator. Its two arms provide architectural variability: the task can be performed using both or one arm.  Each arm has limited range of operation, so when working with one arm, the robot uses the mobility of the Ridgeback platform to reach all the bricks.
\looseness=-1









%







\subsection{Mobile Robot Navigating in a Factory (D2)}
\label{sec:mob_factory}

\noindent
The mobile robot demonstrator (\cref{fig:factory-simulation}) is concerned with navigation on a factory floor, where the robot moves between workstations to perform different manipulation tasks. The robot is the Ridgeback platform, equipped with two Hokuyo lasers, odometry and an inertial measurement unit, and is augmented for this demonstrator with an indoors localization system. The baseline control architecture for navigation is based on the ROS1 navigation stack\,\cite{marder-eppstein.ea:2010}, which uses off-the-shelf control and AI algorithms for planning and reactive obstacle avoidance. The robot navigates to a target location using odometry and sensor data and sending velocity twist commands to the Ridgeback.


The context variability includes:
\begin{enumerate*}[label=(\roman*)]

  \item new obstacles on the factory floor affecting the quality attribute values attained in the default configuration of the navigation stack
  \looseness=-1

  \item variations in power consumption, resulting in changing battery depletion.
  \looseness=-1

\end{enumerate*}
\looseness=-1

%% file: quantitativeevaluation.tex
\section{Feasibility of Runtime Adaptation}%
\label{sec:quantity_evaluation}

\noindent
We analyze the implementation of MROS with respect to its management of component failures and runtime quality attribute levels, by using the two demonstrators and asking the following research questions:

\smallskip

\noindent
\textbf{RQ1.} \emph{How does MROS improve system robustness in the presence of component failures?}

\smallskip

\noindent
\textbf{RQ2.} \textls[-1]{\emph{How does MROS affect system and mission performance in the presence of component failures and system quality concerns?}}

\smallskip

With \textbf{RQ1}, we demonstrate \emph{qualitatively} how MROS addresses component failures on a manipulation mission. \textbf{RQ2} investigates \emph{quantitatively} how the MROS reconfiguration affects the task and system performance of a robot by addressing functional and non-functional concerns in a navigation mission.

\input{study1.tex}

\input{study2.tex}

%% file: study1.tex
\subsection{Manipulation Mission (RQ1)}\label{sec:manipulation}

\subsubsection*{Experiment setup}\label{sec:study1expsetup}

To address \textbf{RQ1}, we conducted a study on D1 dual-arm manipulator (\cref{sec:yumi}) in two scenarios:

\begin{enumerate}

  \item System does not detect the workstation tag.

  \item System cannot use both arms.

\end{enumerate}

\noindent
The first scenario aims to demonstrate the MROS ability to recover a function that requires multiple components (increased mission robustness). The  second scenario demonstrates the MROS controller’s ability to recover from  an  error  by  realising  an  alternative  set  of  functions  to perform  the  task  goal  with  degraded  performance (increased system robustness).

\subsubsection*{Results and discussion}

In the first scenario, the QR-tag for calibration is not detected, due to the light conditions. The meta-controller uses the TOMASys model in \cref{fig:pyramid} to switch between the alternative configurations of the camera parameters \tomasys{FD2.1} and \tomasys{FD2.2} to detect the tag.

In the second scenario, we inject a failure by blocking one of the robot’s arms.  MROS identifies the component error and uses the functional architecture model to find an alternative variant. It adapts the architecture to a configuration that uses the arm not blocked. Since the task cannot be completed with the remaining arm (the wooden blocks are out of range), it also switches the control algorithm to change positions of the mobile platform.
\looseness=-1

\begin{leftbar}

\noindent
We conclude that the MROS meta-control augments the reliability of the overall system, delivering the expected behavior in real-time in the context of component failures.

\end{leftbar}

%% file: study2.tex
\subsection{Navigation Mission (RQ2)}\label{sec:navigation}

\subsubsection*{Experiment setup}

\begin{figure}
\centering
    \begin{subfigure}[b]{0.45\textwidth}
        \centering
        \includegraphics[width=\textwidth]{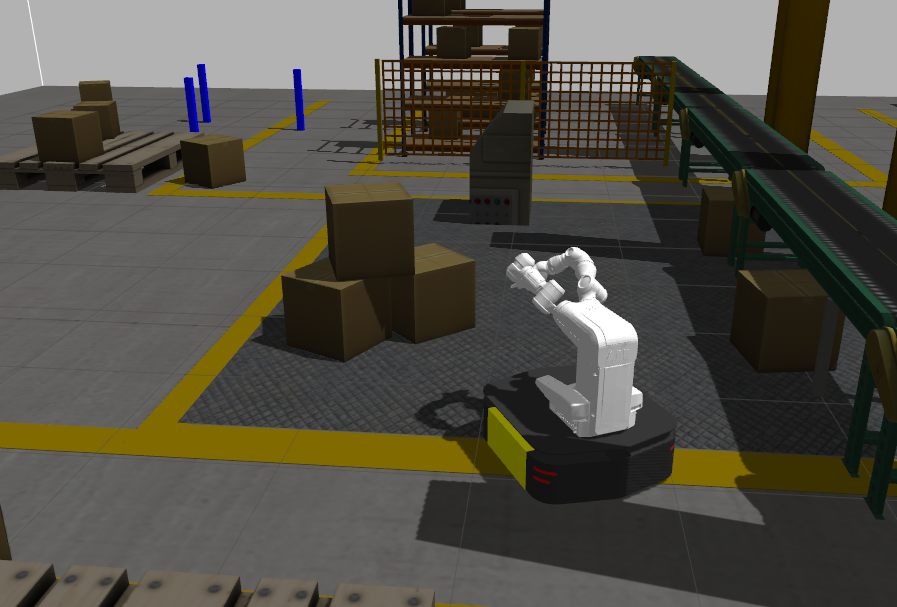}
        \caption{Simulation of the mobile manipulator in a factory.}%
    \label{fig:factory-simulation}
    \end{subfigure}
    \hfill
    \begin{subfigure}[b]{0.45\textwidth}
        \centering
        \includegraphics[width=\textwidth]{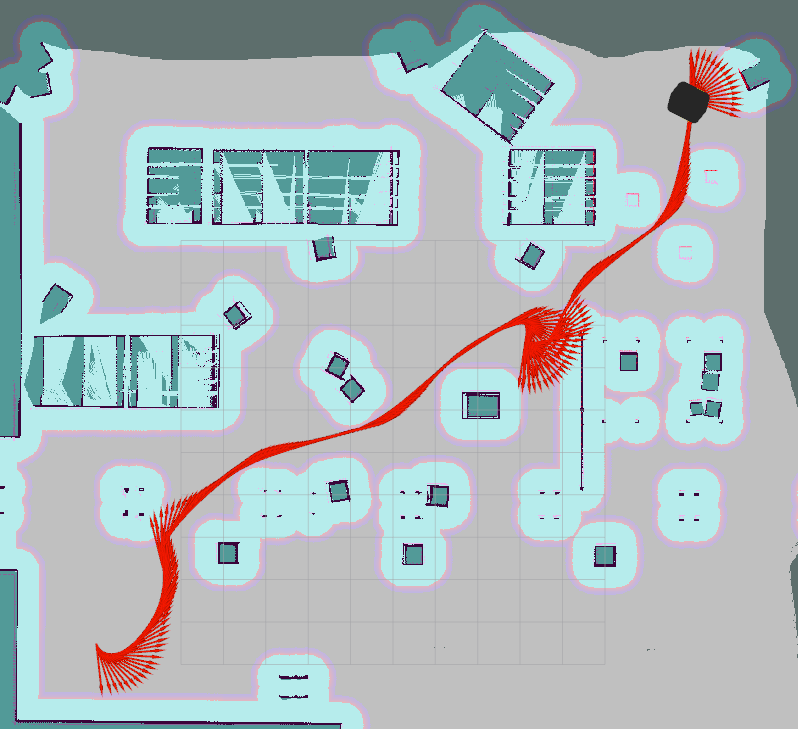}
  \caption{An example route of the mobile manipulator navigating.}%
  \label{fig:nav_mission}
    \end{subfigure}

    \caption{Mobile Manipulator in a factory}
\end{figure}

To address \textbf{RQ2}, we systematically compared a system with the \textit{MROS} meta-controller to a benchmark system (\textit{Baseline}) for mission success, component failure, and quality concerns in the navigation scenario of \cref{sec:mob_factory}.\footnote{The data from this experiment can be found in https://doi.org/10.5281/zenodo.5340886 released under the Creative Commons Attribution 4.0 International license. } The robot is placed at a starting location and it navigates to a goal location. An example of the route during one run is shown in \cref{fig:nav_mission}; the starting position is at the bottom-left of the map and the goal is at the top-right.  We considered six pairs of initial and goal positions.

\subsubsection*{Baseline}

We developed 30 variants of a control architecture, with different quality expected performing the navigation mission. Each of these variants constitutes the \emph{Baseline} system for the corresponding experiment run. The variants are obtained by selecting between a laser or a 3D camera as sensor input for obstacle avoidance, and different values of three parameters that govern the behavior of the \texttt{move\_base} node that generates the motion commands: \texttt{inflation\_radius} (minimal distance allowed to the closest obstacle),  \texttt{max\_vel} (maximum velocity allowed for the robot), \texttt{controller\_frequency} (the frequency of the control loop).

\subsubsection*{MROS-D2}

 In \emph{MROS-D2}, the meta-controller runs on top of one of the architectural variants, performing run-time analysis on component failures and quality violations, and making decisions on when and how to switch to a different navigation configuration.  \emph{MROS-D2} calculates safety at run-time by measuring the braking distance of the robot before hitting an obstacle \texttt{d\_break} \cite{chung2009safe} and the distance from the current location of the robot to the closest obstacle in its direction $P$.  We define runtime safety to be 1 if the robot is at least \texttt{d\_break} distance from the closest obstacle. As the robot approaches the closest obstacle, safety at run-time is computed as $\nicefrac{P}{\text{\texttt{d\_break}}}$.  Furthermore, \emph{MROS-D2} computes the energy at run-time based on an energy model that simulates the power consumption of a robot. The energy model computes the instantaneous \texttt{power\_load} of the robot using current controller frequency, velocity, acceleration, etc. We specify 0.6 as a required threshold for safety and 0.62 for energy. If the robot breaches these thresholds, \emph{MROS-D2} initiates self-adaptation.

The independent variables in our experiment are:
\begin{enumerate*}[label=(\roman*)]

  \item the architectural variant initially deployed for the control architecture, and

  \item the contingencies during the mission:

\end{enumerate*}

\noindent
For each run, we compare the MROS and \emph{Baseline} systems in terms of violation of system qualities thresholds, component failures handling, and mission performance. A run consists of an initial \emph{system configuration C} and a \emph{perturbation P}. We selected nine initial configurations (out of 30) that show the best-expected performance in a set of preliminary tests.  A perturbation models a sequence of contingencies within a mission. Each perturbation $P$ is represented by a tuple $\langle iv_1, iv_2, iv_3\rangle$, where each element represents a value for each contingency type:
\begin{enumerate}

    \item  \emph{Cluttered environment:} the environment contains a set of new obstacles unknown to the robot (not in the map used by the navigation) in randomized positions.  We quantify the level of clutter depending on the number of new obstacles as \emph{low} (7), \emph{medium} (14) and \emph{high} (20).

    \item \textit{Unexpected increase in power consumption:} This could be the result of a sticky surface, a slope, some problem with the wheels, or activation of a new sensor that consumes more power. We quantify this contingency in three levels: 20\%, 40\%, and 60\% increase in power consumption.

    \item \textit{Component failure:} We inject a failure in the laser sensor by killing its driver ROS node at run-time.

\end{enumerate}






 To estimate the effects of MROS on adaptation and mission performance we measure the following dependent variables on both systems:
\begin{itemize}

  \item The \textit{percentage of time the mission is above the threshold for safety risk and energy consumption.} 

  \item \textit{Mission success.} We defined a demanding metric for mission success that combines reaching the goal with performance and safety requirements. A mission is considered successful if (i) the robot reaches its goal, (ii) the average safety risk level for a run is $<0.4$, (iii) the safety risk is below its threshold for more than 95\% of the mission time; (iv) the energy-consumption is below its threshold level for more than 90\% of the time.

  \item \textit{Time to complete} the mission. We measure the time the robot needs to complete a single run.



\end{itemize}

\subsubsection*{Results and Discussion}

\begin{figure}
\centering
    \begin{subfigure}[b]{0.45\textwidth}
        \centering
        \includegraphics[width=\textwidth]{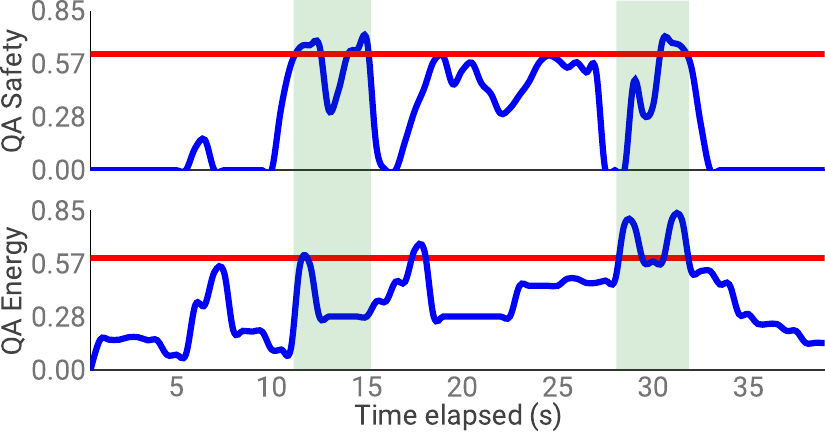}
        \caption{Baseline  without MROS meta-control.}%
        \label{fig:singleRunBase}
    \end{subfigure}
    \hfill
    \begin{subfigure}[b]{0.45\textwidth}
        \centering
        \includegraphics[width=\textwidth]{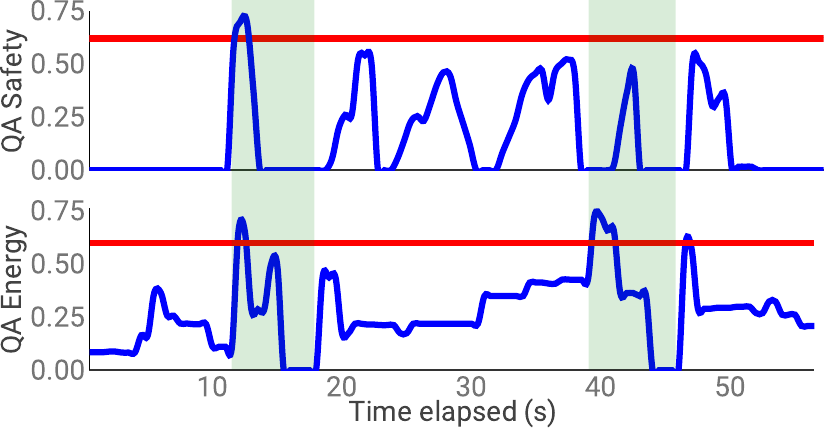}
        \caption{The system with MROS Meta-control.}%
        \label{fig:singleRunMROS}
    \end{subfigure}

    \caption{Safety (top) and Energy (bottom) quality attribute values for a single run. The red line indicates the maximum allowed value in each case. The highlighted areas mark the contingencies where a reconfiguration occurs.}
\end{figure}

We collected data points for 6727 runs of both systems.  \Cref{fig:singleRunBase,fig:singleRunMROS} illustrate the management of safety and energy at run-time during a single run for the Baseline system and for MROS-D2. The red line represents the threshold for each quality, while the highlighted area is a time period when contingencies are introduced.
For the \emph{Baseline} system (\cref{fig:singleRunBase}), the total run-time is 39~seconds, the safety quality attribute value has an average of 0.47 and is above the defined limit for 4~seconds (10\% of the run-time), and the QA Energy has an average of 0.38 and is above the required limit for 3~seconds (8\% of the run-time).
For the \emph{MROS} system (\cref{fig:singleRunMROS}), the total run-time is 56~seconds. In this case, the average of the safety quality attribute value is 0.37 and it is above the defined limit for only 1~second (2\% of the run-time). The average for QA Energy is 0.325 and it is above the required limit for 3~seconds (5\% of the run-time). Moreover, when the MROS system is used, two reconfigurations occur, which can be clearly seen in Figure~\ref{fig:singleRunMROS} where, in the highlighted areas, both QA Energy and Safety values drop down immediately after the limit is crossed and then they restart with a different trend. Each reconfiguration takes about 4.3~seconds, contributing to the MROS run being longer than the \emph{Baseline}.

\begin{table}[t!]

  \renewcommand{\arraystretch}{1.15}

  \begin{tabularx}{\linewidth}{
    X
    r
    r
  }

  \emph{How cluttered the environment is}
  & \emph{MROS}
  & \emph{Baseline}
  \\ \toprule

   no  clutterness contingency
  &	0.99\%
  &	2.60\%
  \\

  low clutterness
  &	0.98\%
  &	2.84\%
  \\

  medium clutterness
  &	0.88\%
  & 1.73\%
  \\

  high clutterness
  &	1.00\%
  &	2.70\%
  \\[1.4mm]

  \textbf{Total}
  &	\textbf{0.96\%}
  & \textbf{2.50\%}
  \\\bottomrule

  \end{tabularx}

 The \caption{Average time a mission is above the safety threshold.}%
  \label{tab:average_sviolation}

  \vspace{-2mm plus 0.5mm minus 0.5mm}

\end{table}

\begin{table}[t!]

  \renewcommand{\arraystretch}{1.15}

  \begin{tabularx}{\linewidth}{
    X
    r
    r
  }

  \emph{Power consumption}
  & \emph{MROS}
  & \emph{Baseline}
  \\ \toprule

  no power contingency
  &	1.26\%
  &	2.63\%
  \\

  power increase 20\%
  &	1.16\%
  &	2.22\%
  \\

  power increase 40\%
  &	2.47\%
  & 2.63\%
  \\

  power increase 60\%
  &	2.63\%
  &	4.57\%
  \\[1.5mm]

  \textbf{Total}
  &	\textbf{1.86\%}
  & \textbf{2.98\%}
  \\ \bottomrule

  \end{tabularx}

  \smallskip

 The \caption{Average time a mission is above the energy threshold.}%
  \label{tab:average_energyviolation}

  \vspace{-2mm plus 0.5mm minus 0.5mm}

\end{table}

\Cref{tab:average_sviolation} shows the average percentage the \emph{Baseline} and \emph{MROS-D2} are above the safety threshold during the mission.
On average \emph{MROS-D2} is 2.6 times more effective in reducing the safety violation from 2.5\% to 0.96\% on average in a single mission.
\emph{MROS-D2} always outperforms the \emph{Baseline}, independently of the level of clutter (number of obstacles). Furthermore, \emph{MROS-D2} manages to keep the safety violation below 1\% for all contingency levels related to a cluttered environment.
Considering these contingency levels, the \emph{MROS-D2} meta-controller re-configures the system once every 2-3 runs.

\Cref{tab:average_energyviolation} shows the average percentage the \emph{Baseline} and \emph{MROS-D2} are above the energy threshold during the mission. Again, \emph{MROS-D2} outperforms the \emph{Baseline} for all contingency cases related to the increase in power consumption. The  biggest  difference  is the  60\%  energy  increase contingency, where  the  result  shows that MROS reduces  the  average violation  of  the  energy constraint  from  4.57\%  to  2.63\%  during the mission.

\begin{table}[t!]

  \renewcommand{\arraystretch}{1.15}

  \begin{tabularx}{\linewidth}{
    X
    r
    r
  }

  \emph{Contingency Type}
  & \emph{MROS}
  & \emph{Baseline}
  \\ \toprule

  no clutter contingency
  &	78.16\%
  & 67.03\%
  \\

  low clutterness
  &	77.92\%
  &	62.03\%
  \\

  medium clutterness
  &	79.13\%
  &	70.16\%
  \\

  high clutterness
  &	77.77\%
  & 62.06\%
  \\[1.5mm]

  no power contingency
  &	81.63\%
  & 59.51\%
  \\

  power increase 20\%
  & 79.38\%
  &	68.48\%
  \\

  power increase 40\%
  &	76.47\%
  & 71.01\%
  \\

  power increase 60\%
  &	76.35\%
  & 60.18\%
  \\[.5mm] \midrule

  component failure
  &	73.00\%
  &	0\%
  \\

  \textbf{Total}
  &	\textbf{78.50\%}
  & \textbf{65.20\%}
  \\ \bottomrule

  \end{tabularx}

  \smallskip

  \caption{Mission success rates under various contingencies.}%
  \label{tab:mission_success}

  \vspace{-2mm plus 0.5mm minus 0.5mm}

\end{table}

\Cref{tab:mission_success} shows that \emph{MROS-D2} outperforms the \emph{Baseline} for all contingency types.  \emph{MROS-D2} improves the mission success by 13.3\% on average. In this calculation, we decided not to include the data for component failure, because the \emph{Baseline} will always fail the mission when a laser failure is introduced (so the 13.3\% reported is biased against \emph{MROS-D2}). In contrast, \emph{MROS-D2} effectively adapts when laser failure is introduced, 
completing the mission in 73\% of the mission runs. Beyond the component failure contingency, the  biggest  difference in mission success can be seen in the case when there is not any increase in power consumption. In this case, the \emph{Baseline} has 59.51\%, while \emph{MROS-D2} has 81.63\% mission success.


Finally, MROS increases the mission completion time by 7.5, as a result a result of the Meta-controller choosing a slower configuration to satisfy the energy and safety quality concerns. As expected in MROS-D2, adaptations were triggered more often for higher contingency levels in terms of clutter and energy peaks.
\looseness=-1

It is important to note that one can engineer a more resilient robot, with more redundant hardware and control strategies, and then the reported numbers for \emph{MROS} will improve further, as they are indeed a function of a more adaptive strategy, which the baseline system does not have.  The experiment demonstrates however that the model-based MROS meta-controller is capable to exploit the hardware and software redundancy to increase robustness. Thus we conclude:

\begin{leftbar}

\noindent Model-based intelligent reconfiguration does improve mission and system performance for robot systems, effectively managing component failures and quality concerns.

\end{leftbar}

%% file: qualitativeevaluation.tex
\section{Qualitative Analysis of MROS Design}%
\label{sec:quality_evaluation}

\noindent
MROS contributes two tools: the \emph{RosModel} MROS plugins for ROS developers at design time, and the Meta-controller ROS packages to deploy it in a ROS system.  In order to understand the generality of our contribution, we discuss the re-usability and extensibility of these artifacts.

\subsection{Reusability}

\noindent
We prototyped the MAPE-K loop of the meta-controller for the D1 dual-arm manipulator demo without generative programming.  Following a common practice in the field, we then generalized it for D2 using model-driven development.

For D1, we designed the OWL implementation of the TOMASys meta-model (\texttt{tomasys.owl}) to support the specification of MROS models, and the ontological reasoning for Steps R2 and R3 (\texttt{mros1\_reasoner.py}).
The TOMASys model was manually developed. We first analyzed the existing software for the system following the \emph{ISE\&PPOOA} MBSE methodology \cite{Fernandez-2019}, obtaining the conceptual representation of its functional architecture and possible variants (\cref{fig:arch-model}). Then we modeled it in OWL, using \texttt{tomasys.owl} and the Protege editor.
For D1, we used application specific solutions for steps R1 and R4 at runtime.
\looseness=-1


To exploit the commonality in the domain, we consequently settled to use the model-driven methods and the \emph{RosModel} bridge tools.  The resulting MROS tools offer:

\begin{itemize}

  \item \emph{Modeling languages and tools for ROS architectures:} (a) linking ROS and TOMASys via \emph{RosSystem} (b) with consistency checking and syntax feedback

  \item \emph{Extracting models from code}, allowing to add meta-control to existing projects and using new nodes with MROS.

  \item \emph{Automatic generation of the ontological knowledge base} (otherwise a tedious and low-level task).

  \item \emph{A reusable monitor of node status} \texttt{rosgraph\_monitor} (used in Step R1)

  \item \emph{A generator of boilerplate code for observers} (step R1).

  \item \emph{A reusable meta-controller}, including tactics to detect and address violations of quality attributes and contingencies (steps R2 and R3)

  \item \emph{A reusable navigation ontology} based on \texttt{tomasys.owl} including minimal domain-specific model of safety, energy and performance quality attributes in navigation, along with the associated SWRL adaptation rules (steps R2, R3).

  \item \emph{Automatic generation of executable reconfiguration actions for the system architecture} (launch files) used by the \textls[-6]{\texttt{rosgraph\_manipulator}} to reconfigure the system (Step\,R4)

\end{itemize}


\looseness=-1



\noindent
For demonstrator D2, we used this infrastructure to model the architectural variants with 30 different \emph{RosSystem} models of the possible configurations of the navigation stack.

It is a well known fact that effective deployments of model-driven engineering require a very good domain implementation.  It is somewhat surprising thus that often model-driven technology is demonstrated in isolation, or on small examples and use cases.  The Robot Operating System provides an excellent platform, a rich and highly customizable implementation of the robotics domain.  ROS with model-driven engineering is a ``match made in heaven.'' Using ROS was an enabling factor for this research: it opened access to standard formats, architectures, and interfaces.  We benefited from the rich tooling and experience ecosystem, which allowed us to efficiently execute rich and realistic experiments, rarely seen in model-based self-adaptation research.  Furthermore, using generative model-driven solutions with ROS meant that the meta-controller technology is readily accessible to unprecedented number engineers and researchers working on self-adaptation in robotics and software engineering using ROS.
\looseness=-1

\subsection{System Extensibility}

\noindent
We analyze how the MROS framework supports system extensibility in the context of a mobile robot navigating in a factory scenario with respect to which aspects should one extend in the application:  (i) to capture new components added to the system; (ii) to capture new quality attributes to be monitored.
\looseness=-1

\subsubsection*{Adding new components}

A 3D camera component was added so the mobile robot would be able to continue the task even if the laser fails. 
To add a new component we only needed to add the new configurations in the \emph{RosSystem} models.
We extended the number of system variants by adding new configurations that contain the camera component.  We used the \emph{RosModel} tools to model 3 new configurations which corresponds to 30 different variants in total.  For monitoring, we did not need to add new observer models, because the current tooling allows component failures to be captured. For analysis and planning, we used the model-to-model transformation to automatically generate the updated knowledge base. Finally, the \emph{RosModel} tools automatically generate the new reconfiguration actions.

\subsubsection*{Adding new quality attributes}

We extended the application scenario D2 with energy-efficiency. For monitoring, we developed an observer for energy-efficiency that calculates energy at run-time.
 Energy level was computed using an energy model that simulates the power consumption of a robot using controller frequency, velocity, acceleration, etc.  For analysis (R2) and planning (R3), we extended the knowledge base with new information about the energy-efficiency of each system configuration. Manual work is unavoidable, but the artifacts are modular and reusable across robotic systems with the same quality attributes.  Again as a final step, the \emph{RosModel} tools automatically generate the reconfiguration actions.
\looseness=-1

In conclusion, in both cases the cost of development of the new features of the robot, greatly outweighs the cost if incorporating it into the meta-control system, which is mostly automatic, barring modeling the new configurations (easy) and new quality attributes (reusable).
\looseness=-1

%% file: related.tex
\section{Related Work}%
\label{sec:related}



\noindent
Dynamic architectures and self-adaptation have been an active topic of research in model-driven software engineering for a decade. The so called \emph{models-at-runtime} paradigm\,\cite{Blair-2009, Bencomo-2019} has been exploring the use of abstractions to reduce complexity and cope with increasingly complex runtime adaptation problems.  One group of prior works related to the use of models in robot software
is based on variability management and feature models that can represent the functionalities provided by a software system symbolically\,\cite{foda,DBLP:conf/vamos/CzarneckiGRSW12}.  
The previous efforts have focused on the solution space of architecture design, that is the possible configurations of components. The discrete variants in the architecture are modeled separately from properties and quality attributes (the problem space). However, formal semantics, a common meta-model, for automated reasoning about the relation between both the solution and the problem space is missing. We address this shortcoming by extending the TOMASys meta-model to represent quality attributes and implement it as an ontology for automated reasoning.
\looseness=-1

Non-functional requirements, such as performance, reliability or safety, have been considered in the dynamic reconfiguration of robot software architecture~\cite{Lotz-2013}. In a newer work, Brugali et al.\ have integrated feature models, standard modelling languages (UML-MARTE) with queuing network models to support reconfiguration in order to maintain an adequate level of performance\,\cite{Brugali-2018}. 
These solutions require intensive modelling efforts, MDE skills and completely new toolsets.  In contrast, we chose a less invasive path, in that we exploit existing codebase and practices in the ROS ecosystem.

Iftikhar and Weyns \cite{iftikhar2014activforms} propose Active FORmal Models for Self-adaptation (ActivFORMS). The adaptation goals that are verified offline are guaranteed to be observed at runtime. The method uses timed automata for modeling the behavior of the multi-robot system, a method unlikely to be accepted by a broader robotics community. Its expressiveness is lower than of linear hybrid control systems, which means they require careful modeling and abstraction for any typical robotics problem.  Our ontology-based approach presented  offers more natural models of trade-offs and requirements, including non-linearities, thanks to the use of description logics.
\looseness=-1

Aldrich et al.\ \cite{aldrich2019model} leverage predictive data models to enable automated robot adaptation to changes in the environment at run-time. While the approach clearly depicts the benefits of using models by capturing high-level artefacts (combining application and system logic), it makes it extremely challenging for a developer to make use of them in robotic scenarios because: (i) it does not introduce models that can be reused for a different application; (ii) it does not give insights how someone can build similar models; (iii) it does not provide automated infrastructure to leverage those models. 

The BRICS\cite{ref_brics} and RobMoSys projects\,\cite{RobMoSysWeb} have proposed model-based  approaches to develop robotic systems. Our work is inspired by the meta-models resulting from those efforts, but goes beyond realizing their proposed future roadmap to use the models at run time for adaptation and autonomy.

Cheng et al. propose \cite{cheng2020ac} a framework that  uses GSN assurance case models to manage run-time adaptations for ROS-based systems. While the framework clearly shows the value of systematically integrating assurance information from GSN models to ROS specific information to guide runtime monitoring and adaptation, it has couple of drawbacks: (i) it uses custom developed libraries specific to the approach,  rather than standard libraries in ROS (such as ROS Diagnostics); (ii) it does not reuse preexisting practises developed in the ROS ecosystem raising the entry barrier for ROS developers to effectively use it. Also, thanks to the tight integration of the ROS framework and model-driven automation, we are able to execute extensive experiments, beyond simple demos. We quantify the improvement of the self-adaptation, and its computational cost.

As discussed in section \ref{sec:background}, our work exploits of two existing streams of contributions. Hernández et al.\,\cite{Hernandez-2018a, Hernandez-2020} proposed principles for knowledge-based self-adaptation and the concept meta-control based on off-the-shelf reasoners, but did not provide a solution to obtain the architectural models or to realize their approach on robotic systems. Hammoudeh et al. \cite{Hammoudeh-2019} created the model-based RosModel tooling to support development ROS systems, but it could not model architectural variants or quality concerns. Our work leverages both works to create a complete solution to develop self-adaptation in ROS systems.
\looseness=-1


%% file: conclusion.tex
\section{Conclusions and Future Work}%
\label{sec:conclusion}

\noindent
This paper introduces the MROS framework to support the integration of self-adaptation in robot control architectures based on ROS.
MROS promotes separation of system management decision making from the mission logic, using ontological reasoning and design models.

MROS uses platform specific DSLs to represent the architectural models of variants of the robotic system. We have developed extensions of the \emph{RosModel} tools that lower the modelling effort for developers and allow to automate the implementation and deployment of the self-adaptation mechanisms.
The use of an ontology-based knowledge based and reasoning for the MROS MAPE-K loop facilitates reusing MROS adaptation rules across robotic applications, and extend them to address new quality concerns beyond safety and energy, which have been addressed in this work.

We developed two MROS tools: \emph{RosModel} tooling extensions to support the development phase, and a ROS package to deploy the MROS meta-controller for run time adaptation, and applied in two different complex robotic systems for validation. Our  results from systematic experiments in the mobile robot demonstrator show how MROS can be used to increase mission reliability and safety and energy concerns. The implementation also illustrates how the MROS DSL approach at design time and ontological approach at run time allowed to easily reuse the adaptation rules from component failure, and extend them to safety concerns and then to energy.

The architecture of MROS provides a framework to explore the integration of methods from robotics, software engineering and artificial intelligence.
For example, currently we are investigating deep learning methods for quality prediction, and the integration of MROS meta-controller with different solutions to manage the mission logic, such as behaviour trees or planning methods.
Moreover, we are working on a version of MROS for ROS2, which offers better support for component and system management than ROS1, and more industrial traction. This will open new opportunities and challenges for self adaptive systems and modelling in robotics.